\documentclass[twocolumn]{article}
\usepackage{framed,multirow}
\usepackage{amssymb}
\usepackage{latexsym}
\usepackage{url}
\usepackage{xcolor}
\definecolor{newcolor}{rgb}{.8,.349,.1}

\usepackage{amsmath}
\usepackage{booktabs}
\usepackage{graphicx}
\usepackage{subcaption}
\usepackage[normalem]{ulem}
\usepackage{dblfloatfix}
\usepackage[noadjust]{cite}
\usepackage{authblk}
\usepackage{hyperref}
\hypersetup{colorlinks,citecolor=black, linkcolor=black, urlcolor=black}	
\usepackage[nameinlink,capitalize]{cleveref}
\usepackage{natbib}

\newcommand{\x}{\mathbf{x}}
\newcommand{\y}{\mathbf{y}}

\newcommand{\itf}{\mathit{f}}

\DeclareMathOperator*{\argmax}{arg\,max~}
\title{Two Demonstrations of the Machine Translation Applications to Historical Documents}
\author{Miguel Domingo and Francisco Casacuberta}
\affil{Pattern Recognition and Human Language Technology Research Center \protect\\ Universitat Polit{\`e}cnica de Val{\`e}ncia - Camino de Vera s/n, 46022 Valencia, Spain \protect\\ $ $ \protect\\ \emph{midobal@prhlt.upv.es, fcn@prhlt.upv.es}}
\date{}

\begin{document}
\maketitle

\begin{abstract}
We present our demonstration of two machine translation applications to historical documents. The first task consists in generating a new version of a historical document, written in the modern version of its original language. The second application is limited to a document's orthography. It adapts the document's spelling to modern standards in order to achieve an orthography consistency and accounting for the lack of spelling conventions. We followed an interactive, adaptive framework that allows the user to introduce corrections to the system's hypothesis. The system reacts to these corrections by generating a new hypothesis that takes them into account. Once the user is satisfied with the system's hypothesis and validates it, the system adapts its model following an online learning strategy. This system is implemented following a client{\textendash}server architecture. We developed a website which communicates with the neural models. All code is open-source and publicly available. The demonstration is hosted at \url{http://demosmt.prhlt.upv.es/mthd/}.
\end{abstract}

\section{Introduction}
\label{se:intro}
Despite being an important part of our cultural heritage, historical documents are mostly accessible to scholars. This is due to both the linguistic properties of these documents{\textemdash}in the past, orthography changed depending on the time period and author due to a lack of an spelling convention{\textemdash}and the nature of language{\textemdash}which evolves with the passage of time. This creates a language barrier which increases the difficulty of comprehending historical documents.

Different research topics on historical documents focus on tackling this language barrier as well as different aspects related to the document's language richness, which is also part of our cultural heritage. For example, historical manuscripts are automatically digitized and transcribed \citep{Toselli17}. Documents are automatically translated into a modern version of their original language to make them accessible to a broader audience \cite{Domingo20a}. Orthography is normalized to account for the lack of a spelling convention \citep{Porta13}. Search queries can find all occurrences of one or more words \citep{Ernst06}. Word frequency lists are generated \citep{Baron09}. And natural language processing tools provide automatic annotations to identify and extract linguistic structures such as relative clauses \citep{Hundt11} or verb phrases \citep{Pettersson13}.

In this work, we focus on two of these topics: language modernization and spelling normalization. The first one aims to generate a modern version of a historical document to make its content available to a broader audience. The second one aims to achieve an orthography consistency{\textemdash}without altering the document's content{\textemdash}to reduce the variability derived from the lack of spelling conventions and facilitate the work of scholars. We present a demonstration that showcases neural machine translation (NMT) applications to these two topics, and provides an interactive, adaptive framework for scholars to increase their productivity when working on these tasks.

\section{Interactive, adaptive NMT framework}
\label{se:MT}
Given a source sentence $\x$, machine translation (MT) aims to find the most likely translation $\hat{\y}$ \citep{Brown93}:

\begin{equation}
\hat{\y} = \argmax_{\y} Pr(\y \mid \x)
\label{eq:smt}
\end{equation}

NMT models this equation with a neural network which usually follows an encoder-decoder architecture (although others architectures are possible), in which the source sentence is projected into a distributed representation at the encoding step. Then, at the decoding step, the decoder generates, word by word, its most likely translation using a beam search method \citep{Sutskever14}. The model parameters are typically estimated jointly on large parallel corpora, via stochastic gradient descent \citep{Robbins51,Rumelhart86}. At decoding time, the system obtains the most likely translation by means of a beam search method.

Despite the quality improvement in recent years, NMT is still far from obtaining high-quality translations \citep{Toral18}. Thus, human experts need to supervise and correct the NMT systems outputs in order to achieve the highest quality. As an alternative to this static, decoupled strategy for correcting systems, protocols such as the iterative-predictive have been studied for years \citep{Barrachina09,Knowles16,Peris19}.

In this framework, the correction process shifts to a human{\textendash}computer collaboration. The user interacts with the system through a feedback signal $\itf$ (e.g., a word correction). Then, the system suggests a new hypothesis $\tilde{\y}$, compatible with the user's feedback. This restricts \cref{eq:smt} by constraining the search space:

\begin{equation}
\tilde{\y} = \argmax_{\y \mathrm{\ compatible\ with\ } \itf} Pr(\y \mid \x, \itf)
\label{eq:imt}
\end{equation}

This interactive correction process is repeated until the user is satisfied with the system's suggestion. This validated translation is new valuable data that could be leveraged to continuously improve the system. Each time the user validates a translation hypothesis, the system's models can be incrementally updated with this new sample. Therefore, when the system generates a new translation, it will consider the previous user feedback and it is expected to produce higher quality translations. A common way of generating this adaptive systems consists in following an online learning paradigm \citep{Ortiz16,Peris19}.

\section{Applications}
\label{se:tasks}
Our demonstration showcases NMT applications to two different historical documents research topics: language modernization and spelling normalization. 

\subsection{Language modernization}
This research topic aims to tackle the language barrier inherent in historical documents in order to make them available to a broader audience. To do so, it proposes to automatically generate a new version of a historical document, written in the modern version of the document's original language. A common approach to this problem is to tackle modernization as a conventional MT task \citep{Domingo17a,Sen19}. Results showed that, while there is still room for improvement, modernization techniques successfully decrease the comprehension difficulty of historical documents{\textemdash}indicated by both automatic metrics and human evaluation \citep{Domingo20a}.

Additionally, while modernization's goal is limited to bringing a better understanding of historical documents to a general audience{\textemdash}language-related losses may appear during the process{\textemdash}scholars are in charge of different tasks that require them to generate modernizations of the highest quality (e.g., producing a comprehensive contents document for non-experts \citep{Monk18}). Thus, we provided our system of an interactive, adaptive framework to increase the scholars productivity when working on these tasks. The modernization system was developed following \citet{Domingo20a}. For the interactive, adaptive framework, we followed \citet{Peris19}. 

\subsection{Spelling normalization}
In order to account for the lack of spelling conventions, which were not created until recent years, spelling normalization aims to achieve an orthography consistency by adapting a document's spelling to modern standards. This linguistics problem suppose an additional challenge for the effective natural language processing for these documents. Through the years, different approaches to this problem have been researched \citep[e.g.,][]{Baron08,Porta13}. Some of them tackle spelling normalization as a conventional MT problem \citep[e.g.,][]{Scherrer13,Bollman18}. We developed our normalization system following \citet{Domingo19c}, which approached spelling normalization using character-based NMT and obtained significant improvements according to several automatic metrics. Like in the previous task, we followed \citet{Peris19} for the interactive, adaptive framework.

\section{System description}
\label{se:system}
Our system is composed of two main elements: the client and the server. The client is an HTML website, which interacts with the user through javascript and communicates with the server via the HTTP protocol, using the PHP curl tool. The server is the core element. It contains the NMT systems, which were developed with \emph{NMT-Keras} \citep{Peris18}, and it is deployed as a Python HTTP server that handles the client's requests. All code is open-source an publicly available\footnote{\url{https://github.com/midobal/mthd}.}.

Initially, an old sentence is presented to the user in the client website. When the user requests an automatic modernization/normalization (see \cref{se:tasks}), the client communicates the server via PHP. Then, the server queries the NMT system, which generates an initial hypothesis applying \cref{eq:smt}. After that, the hypothesis is sent back to the client website.

\begin{figure}[!ht]
	\centering
	\includegraphics[scale=0.5]{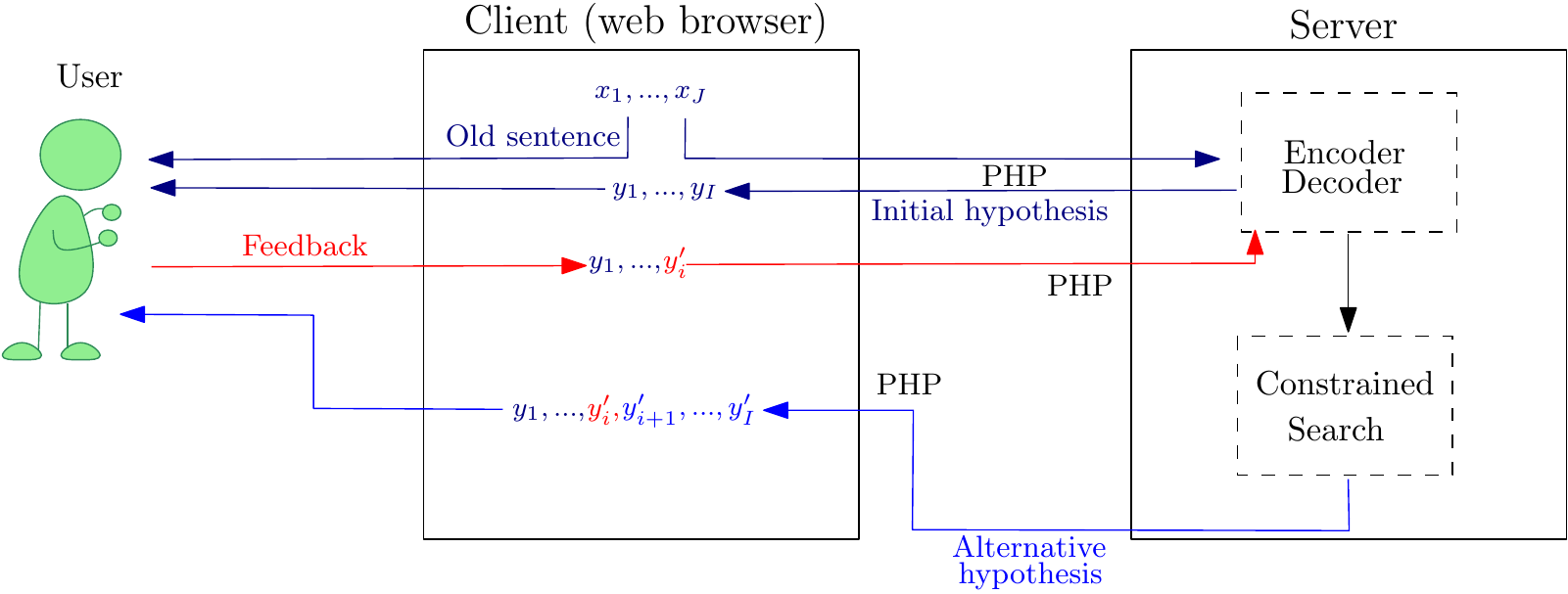}
	\caption{System architecture. The client presents the user an old sentence and a prediction. Then, the user introduces a feedback signal for correcting this prediction (in this example, they are validating the prefix \emph{\small $y_1,..,y_{i-1}$} and correcting the word \emph{\small $y^{\prime}_i$}). After that, the old sentence and the user's feedback is sent to the server, which generates an alternative hypothesis that takes into account the user corrections (in this example, a new suffix \emph{\small $y^{\prime}_{i+1},..,y^{\prime}_I$} that completes the user's feedback).}
	\label{fi:architecture}
\end{figure}

At this point, the interactive-predictive process starts: The user searches the hypothesis for the first error and introduces a correction with the keyboard (writing one or more characters). Once the user finishes typing, the client reacts to this feedback by sending a request to the server. This request contains the old sentence and the user feedback (the sequence of characters that conform the prefix). Then, the NMT system applies \cref{eq:imt} to produce an alternative hypothesis coherent with the user's feedback and sends it back to the client website. This process is repeated until the user finds the system's hypothesis satisfactory. \cref{fi:architecture} illustrates one step of this process.

\begin{figure*}[!ht]
	\centering
	\centerline{\includegraphics[scale=0.22]{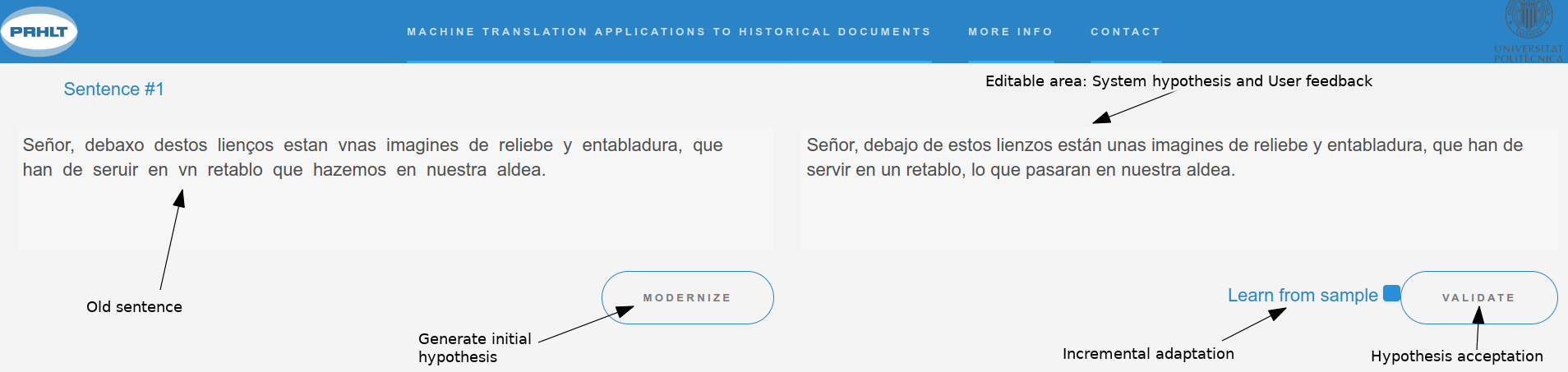}}
	\caption{Frontend of the client website. As the button ``Modernize'' is clicked (or ``Normalize'', depending on the task your are performing), an initial hypothesis for the old sentence appears in the right area. Then, the user can introduce corrections of this text. The system will react to each correction, producing alternative hypotheses coherent with the user feedback. Once the user is satisfied with the modernization hypothesis, they can click in the ``Validate'' button to accept the hypothesis.}
	\label{fi:example}
\end{figure*}

Once the user is satisfied with the system's hypothesis, they can validate it. Then, the system is incrementally updated with this new sample following an online learning setup \citep{Peris19}. Hence, in future interactions, the system will be progressively updated and able to generate better hypothesis.

\cref{fi:example} illustrates an example of how to performed a task using the client server. After having selected the task to perform, a list of old sentences will appear. When you click on ``Modernize/Normalize'', the system will generate an initial hypothesis. If you desire to improve this hypothesis, you can click on the left box and type a correction. The system will, then, generate a new hypothesis to take that correction into account. You can repeat this process for as many corrections as you desire to make. Finally, you can click ``Validate'' to tell the system that you are happy with the modernization/normalization and, if the \emph{Learn from sample} option is activated (in blue), the system will use the sample to improve its model.

\section{Conclusions and future work}
\label{se:conc}
We presented a demonstration of two MT applications to historical documents. We described its client{\textendash}server architecture and developed a website to facilitate the use of the system.

As a future work, we would like to improve our website's front end. This improvement could allow us to create a better visualization of the hypothesis, which is specially relevant for the spelling normalization task. Additionally, relevant attributes of the neural system could be visualize in order to  help to understand better the model predictions and behavior.

\section*{Acknowledgments}
The research leading to these results has received funding from Generalitat  Valenciana  (GVA)  under  project PROMETEO/2019/121.   We gratefully  acknowledge  the  support  of  NVIDIA  Corporation  with the donation of a GPU used for part of this research, and Andr{\'e}s Trapiello and Ediciones Destino for granting us permission to use their book in our research.

\bibliographystyle{apalike}
\bibliography{mthd}

\end{document}